\documentclass[10pt,twocolumn,letterpaper]{article}

\usepackage[pagenumbers]{cvpr}

\usepackage{graphicx}
\usepackage{amsmath}
\usepackage{amssymb}
\usepackage{booktabs}
\usepackage{color}
\usepackage{multirow}
\usepackage{bbm}
\usepackage[table]{xcolor}
\usepackage{arydshln}
\usepackage{algorithm}
\usepackage{listings}
\usepackage{multicol}
\usepackage[T1]{fontenc}

\usepackage{txfonts}

\definecolor{mygreen}{HTML}{39b54a}  
\definecolor{myred}{HTML}{A10035}
\definecolor{myyellow}{HTML}{F8E924}
\definecolor{ggray}{RGB}{127,127,127}
\definecolor{mygray1}{gray}{.5}
\definecolor{mygray}{gray}{.9}
\definecolor{aliceblue}{rgb}{0.94, 0.97, 1.0}

\usepackage{etoolbox}
\makeatletter
\AfterEndEnvironment{algorithm}{\let\@algcomment\relax}
\AtEndEnvironment{algorithm}{\hrule\relax\vskip3pt\@algcomment}
\let\@algcomment\relax
\newcommand\algcomment[1]{\def\@algcomment{\footnotesize#1}}
\renewcommand\fs@ruled{\def\@fs@cfont{\bfseries}\let\@fs@capt\floatc@ruled
  \def\@fs@pre{\hrule height.8pt depth0pt}%
  \def\@fs@post{}%
  \def\@fs@mid{\hrule}%
  \let\@fs@iftopcapt\iftrue}
\makeatother

\definecolor{voc_cow}{HTML}{0C1E7F}
\definecolor{voc_horse}{HTML}{FB2576}

\newcommand{\pub}[1]{\color{gray}{\tiny{[{#1}]}}}

\newcommand{\baseline}[1]{\color{ggray}{\scriptsize{{#1}}}}
\newcommand{\tablestyle}[2]{\setlength{\tabcolsep}{#1}\renewcommand{\arraystretch}{#2}\centering\footnotesize}
\newcommand{\bslimp}[2]{
	\textbf{#1} \fontsize{7.5pt}{1em}\selectfont\color{purple}{$\uparrow$ \textbf{#2}}
}

\usepackage{tabulary}
\newcolumntype{I}{!{\vrule width 1pt}}

\newcolumntype{x}[1]{>{\centering\arraybackslash}p{#1pt}}
\newcolumntype{y}[1]{>{\raggedright\arraybackslash}p{#1pt}}
\newcolumntype{z}[1]{>{\raggedleft\arraybackslash}p{#1pt}}
\newlength\savewidth

\makeatletter
\newcommand{\thickhline}{%
	\noalign {\ifnum 0=`}\fi \hrule height 1pt
	\futurelet \reserved@a \@xhline
}
\makeatother

\usepackage[pagebackref,breaklinks,colorlinks]{hyperref}

\usepackage[capitalize]{cleveref}
\crefname{section}{Sec.}{Secs.}
\Crefname{section}{Section}{Sections}
\Crefname{table}{Table}{Tables}
\crefname{table}{Tab.}{Tabs.}

\usepackage[misc]{ifsym}

\begin{document}

\title{Out-of-Candidate Rectification for Weakly Supervised Semantic Segmentation}

\author{
Zesen Cheng$^{1}$\thanks{Equal~contribution.} \quad Pengcheng Qiao$^{1*}$ \quad Kehan Li$^{1}$ \quad Siheng Li$^{3}$ \quad Pengxu Wei$^{4}$ \\
Xiangyang Ji$^{3}$ \quad Li Yuan$^{1,2}$ \quad Chang Liu$^{3}$~\textsuperscript{\Letter} \quad Jie Chen$^{1,2}$~\textsuperscript{\Letter} \and
$^{1}$ School of Electronic and Computer Engineering, Peking University \\
$^{2}$ Peng Cheng Laboratory \quad
$^{3}$ Tsinghua University \quad 
$^{4}$ Sun Yat-Sen University \quad \\
}

\maketitle

\begin{abstract}
Weakly supervised semantic segmentation is typically inspired by class activation maps, which serve as pseudo masks with class-discriminative regions highlighted.
Although tremendous efforts have been made to recall precise and complete locations for each class, existing methods still commonly suffer from the unsolicited \textbf{Out-of-Candidate} (OC) error predictions that not belongs to the label candidates, which could be avoidable since the contradiction with image-level class tags is easy to be detected.
In this paper, we develop a group ranking-based \underline{\textbf{O}}ut-of-\underline{\textbf{C}}andidate \underline{\textbf{R}}ectification (OCR) mechanism in a plug-and-play fashion. 
Firstly, we adaptively split the semantic categories into \textbf{In-Candidate} (IC) and OC groups for each OC pixel according to their prior annotation correlation and posterior prediction correlation.
Then, we derive a differentiable rectification loss to force OC pixels to shift to the IC group.
Incorporating our OCR with seminal baselines (e.g., AffinityNet, SEAM, MCTformer), we can achieve remarkable performance gains on both Pascal VOC (+3.2\%, +3.3\%, +0.8\% mIoU) and MS COCO (+1.0\%, +1.3\%, +0.5\% mIoU) datasets with negligible extra training overhead, which justifies the effectiveness and generality of our OCR.
Code is available at \rm{\url{https://github.com/sennnnn/Out-of-Candidate-Rectification}}.
\end{abstract}


\begin{figure}[t]
\centering
\includegraphics[width=0.48\textwidth]{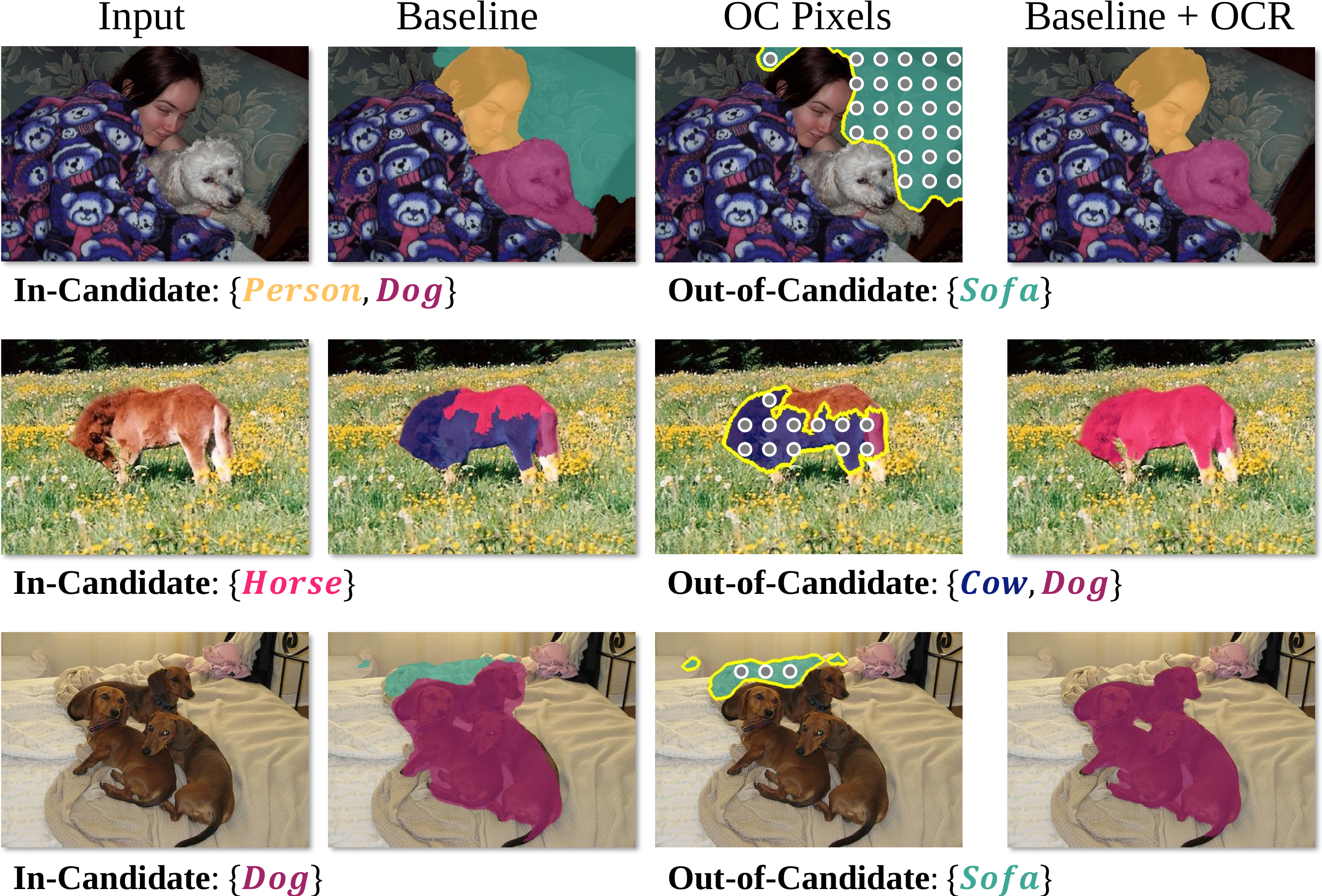}
\caption{\textbf{Motivation of our OCR}. We visualize the segmentation results from the baseline method (e.g. SEAM) and the baseline with our proposed OCR. The predictions from baseline methods are easily disturbed by OC pixels, that is, pixels whose semantic categories are in contradiction with label candidate set (inner of the \textbf{\color{myyellow}{Yellow}} contour). Our proposed OCR can rectify these OC pixels and suppress this unreasonable phenomenon.}
\label{fig:phenomenon}
\vspace{-15pt}
\end{figure}
\begin{figure*}[t]
\centering
\includegraphics[width=0.95\textwidth]{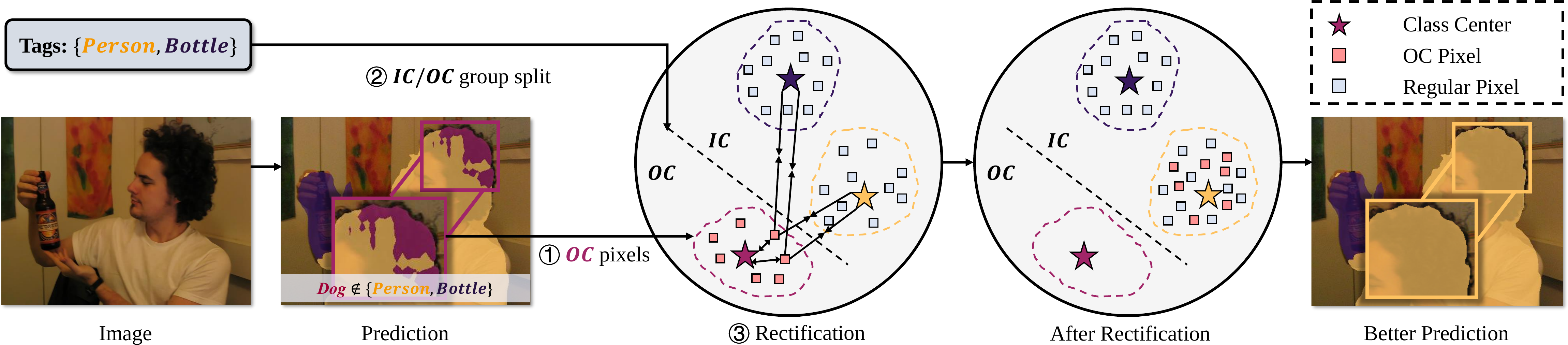}
\caption{\textbf{Conceptual workflow of our OCR}. The OC pixels are selected out by checking if the semantic categories are in contradiction with image-level candidate tags. Then we adaptively split the categories into IC group and OC group. Finally, we utilize rectification loss for group ranking and let OC pixels escape from OC group to IC group.}
\label{fig:intuition}
\vspace{-10pt}
\end{figure*}

\section{Introduction}
\label{sec:intro}

Due to the development of deep learning, significant progress has been made in deep learning-based semantic segmentation\cite{long2015fully,ronneberger2015u}. 
However, its effectiveness requires huge amounts of data with precise pixel-level labels. Collecting precise pixel-level labels is very time-consuming and labor-intensive, thus much research shifts attention to training effective semantic segmentation models with relatively low manual annotation cost, i.e., Weakly Supervised Semantic Segmentation~(WSSS). 
There exist various types of weak supervision for semantic segmentation such as image-level tag labels~\cite{hong2017weakly,ahn2018learning,ahn2019weakly,wang2020self,zhang2020splitting}, bounding boxes~\cite{khoreva2017simple,dai2015boxsup,lee2021bbam}, scribbles~\cite{lin2016scribblesup,vernaza2017learning} and points~\cite{bearman2016s}. In this work, we focus on WSSS based on image-level tag labels since image-level tags demand the least annotation cost, which just needs the information on the existence of the target object categories.

Most of the previous WSSS methods follow such a standard workflow~\cite{ahn2018learning}: 1). generating high-quality Class Activation Maps~(CAM)~\cite{wang2020self,xu2022multi}; 2). generating pseudo labels from CAMs~\cite{ahn2018learning,zhang2021adaptive}; 3). training segmentation networks from pseudo labels. 
Previous works mainly focus on the first and second procedures.
However, training segmentation network from pseudo labels is also vital because neural network can exploit shared patterns between pseudo labels~\cite{arpit2017closer} and largely improve final segmentation results~\cite{li2021pseudo}.
But the pseudo label generation relies on high-quality CAM, while the pseudo labels usually are incomplete and imprecise because CAM only focuses on discriminative object parts and can not fully exploit object regions~\cite{chang2020weakly}.
The existence of noise in pseudo labels provides confused knowledge to segmentation networks and results in error predictions. According to the observations in Fig.~\ref{fig:phenomenon}, the segmentation networks trained by noisy pseudo labels usually output pixels with semantic categories that do not belong to the candidate label set, i.e., image-level tag labels. This special type of prediction errors are defined as \textbf{\textit{Out-of-Candidate}}~(OC). These errors can be easily detected by checking if the semantic category of pixel is in contradiction with image-level tag labels, which is seldom considered before. For better identifying this phenomenon, we extra name these error pixels as OC pixels and name the illegal categories as OC categories. In contrast, the potentially correct categories for OC pixels are defined as \textbf{\textit{In-Candidate}}~(IC) categories.

To suppress the occurrence of OC phenomenon, we propose group ranking-based \underline{\textbf{O}}ut-of-\underline{\textbf{C}}andidate \underline{\textbf{R}}ectification~(OCR) to rectify OC pixels from OC categories to IC categories by solving a group ranking problem (i.e., the prediction score of IC group needs to be larger than the prediction score of OC group).
In Fig.~\ref{fig:intuition}, OCR is illustrated as three procedures: OC pixels selection, IC/OC categories group split and rectification. 
Firstly, we find out OC pixels whose classification result is in contradiction with image-level tag labels. 
Secondly, we adaptively split the classes into IC classes group and OC classes group for each OC pixel by considering prior label correlation information from the image-level tag labels and posterior label correlation information from the network prediction. 
Finally, rectification loss is used to modulate the distance between OC pixels and class centers of IC group and OC group. It constraints that the OC pixels and OC class centers are pushed away and the OC pixels and IC class centers are pulled closer so that those OC pixels are rectified to correct classes.

Out-of-Candidate Rectification~(OCR) is designed in a plug-and-play style to provide reasonable supervision signals with trivial training costs and to improve evaluation results with no extra cost for inference. To fairly show the effectiveness and generality, we adopt the same settings of several previous methods (AffinityNet~\cite{ahn2018learning}, SEAM~\cite{wang2020self}, MCTformer~\cite{xu2022multi}) and evaluate our proposed OCR on the PASCAL VOC 2012 and MS COCO 2014 datasets. Experiments demonstrate that our OCR improves the performance of final segmentation results. Specifically, our module improves AffinityNet, SEAM and MCTformer by 3.2\%, 3.3\% and 0.8\% mIoU on PASCAL VOC 2012 dataset and 1.0\%, 1.3\% and 0.5\% mIoU on MS COCO 2014 dataset. 

\section{Related works}
\label{sec:related}

Most current WSSS approaches are built upon such a paradigm:~1).~Generating high-quality CAMs; 2).~Refining CAM for pseudo label generation; 3).~Training segmentation network from pseudo labels. The core technology of this paradigm is CAMs~\cite{zhou2016learning}. However, the raw CAMs can only cover the discriminative part of object regions and are thus unable to provide supervision without noise for learning semantic segmentation networks. Previous methods propose their schemes for improving three procedures of the paradigm for easing the limitations:

\par\textbf{Generating high-quality CAMs}
How to generate high-quality CAMs is the key research topic of WSSS because the improvement of CAMs can directly boost the whole workflow from the source. The main purpose of this type of research is to let CAMs cover object regions as precisely and completely as possible. A few methods design heuristic strategies, like "Hide \& Seek"~\cite{singh2017hide} and Erasing~\cite{wei2017object}, adopted on images~\cite{zhang2021complementary, li2018tell,yoon2022adversarial} or feature maps~\cite{lee2019ficklenet, hou2018self,chen2022class} to force the network to exploit novel regions rather than only discriminative regions. Besides, there are also other strategies, like utilizing sub-categories~\cite{chang2020weakly}, self-supervised learning~\cite{shimoda2019self,wang2020self,chen2022self}, contrastive learning~\cite{du2022weakly,xie2022contrastive,ke2021universal,zhou2022regional} and cross-image information~\cite{fan2020cian, sun2020mining, li2021group} to generate precise and complete CAMs. Recently, Vision Transformer (ViT)~\cite{dosovitskiy2020image} is proposed as a new generation of visual neural networks. Because of the long-range context nature, vision transformer can better capture semantic context so some recent works begin to utilize ViT as the classification network for generating high-quality CAMs~\cite{gao2021ts,xu2022multi,ru2022learning,rossetti2022max}. What's more, some of previous methods try to improve the local receptive field of classification network~\cite{wei2018revisiting, xu2021atrous,jiang2022l2g}.  Improving CAMs by modifying the standard classification loss is also researched by previous works, e.g., ~\cite{zhang2020splitting,wu2022adaptive}.

\par\textbf{Generating pseudo labels from CAMs.} 
After acquiring high-quality CAMs, we decode them into pseudo labels. Originated from AffinityNet~\cite{ahn2018learning}, some works focus on exploiting pixel-level affinity learning which can facilitate the generation of higher-quality pseudo labels from CAMs~\cite{ahn2019weakly,wang2020weakly,fan2020cian,zhang2021adaptive,xu2021leveraging}. Besides, prior other works design diverse mechanisms for pseudo label generation, for instance, densecrf~\cite{zhang2019reliability}, texture exploiting~\cite{araslanov2020single} and multi-estimations~\cite{fan2020employing} for better pseudo label generation.

\par\textbf{Training from pseudo labels.}
Although dozens of advanced CAM generation or refinement technology are proposed, the existence of noise in pseudo labels is inevitable. In this context, some works explore about how to train segmentation network from pseudo labels better? For example, URN~\cite{li2022uncertainty} use uncertainty estimation to mitigate noise of pseudo labels when training segmentation networks. Different from previous methods, our works mainly focus on solving OC pixels which are a special type of prediction error firstly defined by us.

\begin{figure*}[t]
\centering
\includegraphics[width=0.96\textwidth]{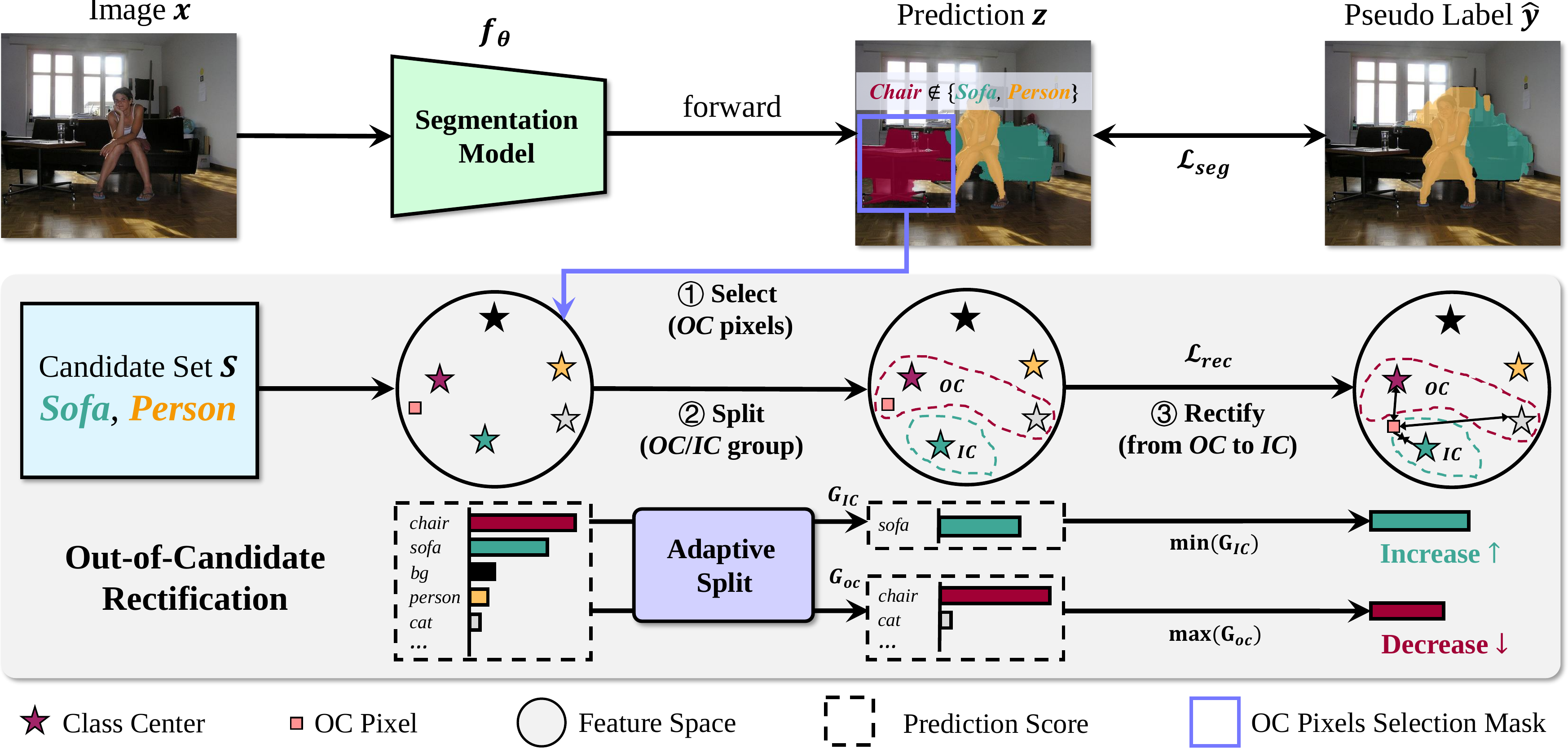}
\caption{\textbf{Detailed workflow of our OCR} which illustrates the behavior about how to rectify an OC pixel from OC classes group to IC classes group. Note that IC classes group is just the candidate set of correct class for single pixel. So, we adopt an adaptive strategy to filter out useless classes in IC classes group when splitting classes into IC classes group and OC classes group, which can reduce the probability of containing incorrect classes in IC classes group. For instance, both "\textit{sofa}" and "\textit{person}" belong to tag labels. However, the "\textit{sofa}" is the correct class rather than "\textit{person}" and involving "\textit{person}" in IC classes group is not beneficial to rectify OC pixels to correct class. The adaptive split strategy filter out "\textit{person}" class according to the correlation information for correct rectification.}
\label{fig:framework}
\vspace{-10pt}
\end{figure*}

\section{Method}
\label{sec:method}

\subsection{Preliminaries}
\label{subsec:pre}

We first define WSSS using the following setup. Let $ \mathcal{X} $ be the input image space, and $ \mathcal{Y} = \{1, 2, ..., C\} $ be the output image-level label space. After sampling from the input image space and output label space, we can define a training dataset $ \mathcal{D} = \{(x_i, S_i)\}^{n}_{i=1} $, where each tuple comprises an image $ x_i \in \mathcal{X} $ and an image-level label set $ S_i \subset \mathcal{Y} $. Equivalent to the supervised semantic segmentation setup, the goal of WSSS is to obtain a pixel classifier that can identify the real class of each pixel. The main difference is that the WSSS setup can not be accessible to exact pixel-level labels. This limitation lets segmentation algorithms struggle to acquire correct pixel-level supervision signals without interference. A basic assumption of WSSS is that the ground truth label of pixels $ y_i^{(x,y)} $ belongs to image $ x_i $ is concealed in label set $ S_i $, i.e., $ y_i^{(x,y)} \in S_i $, but it's invisible to learning models. According to this assumption, previous methods for WSSS leverage class activation map (CAM)~\cite{zhou2016learning,gao2021ts} to generate pseudo segmentation labels $ \hat{y}_i = \{\hat{y}_i^{(x,y)}|\hat{y}_i^{(x,y)}\in S_i\}^{H\times W}_{x=1,y=1} $ which are a type of noisy estimation of real labels $ y_i $. With these pseudo segmentation labels, we can define an extra segmentation noisy dataset $ D_s = \{(x_i, \hat{y}_i)\}^n_{i=1} $. Then a segmentation network $ f_\theta $ is used to fit segmentation noisy dataset $ D_s $ for generating final segmentation results $ z_i $.

\subsection{Overall Pipeline}

To prove the generality of our approach, we build our pipeline upon several previous baseline methods: AffinityNet \cite{ahn2018learning}, SEAM \cite{wang2020self} and MCTformer \cite{xu2022multi}. These methods provide a standard workflow: 1). Generating CAMs; 2). Refining CAM with affinity learning; 3). Training segmentation network.

\par\textbf{Generating CAMs.} For acquiring CAMs, we first pretrain a multi-label classifier $f^{'}_{\theta} $. Then we use a multi-label classification loss $ \mathcal{L}_{cls} $ to supervise the classifier:
\begin{equation}
\begin{split}
    \mathcal{L}_{cls} = E_{(x_i, S_i)\sim D}\bigg[\sum^{C}_{j=1}&-S_i^{j}\mathrm{log}(\sigma(h_i^{j})) \\&- (1 - S_i^{j})\mathrm{log}(1 - \sigma(h_i^{j}) \bigg],\!
\end{split}
\end{equation}
where $ \sigma(\cdot) $ is sigmoid function, $ h^{j}_i $ is the classification logits of $j$-th category and $i$-th sample, i.e., $ h^{j}_{i} = \mathrm{GAP}(f^{'}_{\theta}(x_i)) $. $ f^{'}_{\theta}(x_i) \in \mathbb{R}^{C\times H \times W}$. $ \mathrm{GAP} $ is the global average pooling operation. After pretraining, we need to normalize the initial CAMs to get "seed CAM" $ \mathcal{P} $:
\begin{equation}
\begin{split}
     \mathcal{P}^j_{i} = \frac{f^{'}_{\theta}(x_i)^{j}}{\mathrm{max}(f^{'}_{\theta}(x_i)^{j})}.
\end{split}
\end{equation}

\par\textbf{Refining CAMs with affinity learning.}
Ahn et al.~\cite{ahn2018learning} propose AffinityNet to learn the affinities between adjacent pixels from the reliable seeds of "seed CAM".
Then the AffinityNet is used to predict an affinity matrix to refine "seed CAM" to pseudo mask labels by random walk. Following the settings of baselines, we adopt this refinement.

\par\textbf{Training segmentation network.} Following the settings of baseline methods \cite{ahn2018learning,wang2020self,xu2022multi} and some previous works \cite{zhang2021complementary,xu2021leveraging,zhang2019reliability}, we choose DeepLabv1 \cite{chen2014semantic} based on ResNet38 \cite{wu2019wider} as the segmentation network $ f_{\theta} $. The segmentation network is used to fit pseudo segmentation labels $ \hat{y} $ and the training objective is below:
\begin{equation}
    \mathcal{\mathcal{L}}_{seg} = E_{(x_i, \hat{y}_i)\sim D_s}\left[\sum^{C}_{j=1}-\hat{y}_i^{j}log(\frac{exp(z_i^{j})}{\sum^{C}_{k=1}exp(z_i^{k})})\right]\!\!,
\end{equation}
where $ z_i $ is its logits output, i.e., $ f_{\theta}(x_i) = z_i $. In order to make the formula more clear, we ignore the pixel location index $ (x, y) $ and sample index $ i $. Our method modify the standard training objective and introduce an extra loss:
\begin{equation}
    \mathcal{L} = \mathcal{L}_{seg} + \alpha \mathcal{L}_{rec},
\end{equation}
where $ \mathcal{L}_{rec} $ is the rectification loss which will be introduced below and $ \alpha $ is the loss modulation coefficient.

\subsection{Out-of-Candidate Rectification}
\label{subsec:ocr}

To our best knowledge, previous methods ignore the OC problem during training segmentation network from pseudo segmentation labels. We are the first to design relative mechanism~(OCR) to suppress the occurrence of OC pixels. 
The proposed OCR is comprised of three parts: OC pixels selection, IC and OC group split and rectification loss. The three steps are introduced below.

\textbf{OC Pixels Selection.} When training segmentation network, previous methods only utilize the pseudo segmentation labels $ \hat{y} $. We find that the prior information provided by candidate label set $ S $ is critical. With the prior information, those OC pixels can be easily detected. We implement this detection by designing a mask $ m_{oc} $ (for a single pixel):
\begin{equation}
m_{oc}=\left\{
\begin{aligned}
1 & , & & \mathop{\mathrm{argmax}}\limits_{k}(z^{k})\in \overline{S}, \\
0 & , & & \mathop{\mathrm{argmax}}\limits_{k}(z^{k})\in S \cup \{bg\},
\end{aligned}
\right.
\end{equation}
where $ z^k $ is the segmentation logits of $i$-th class and $ bg $ is the background class.

\begin{figure}[t]
\centering
\includegraphics[width=0.48\textwidth]{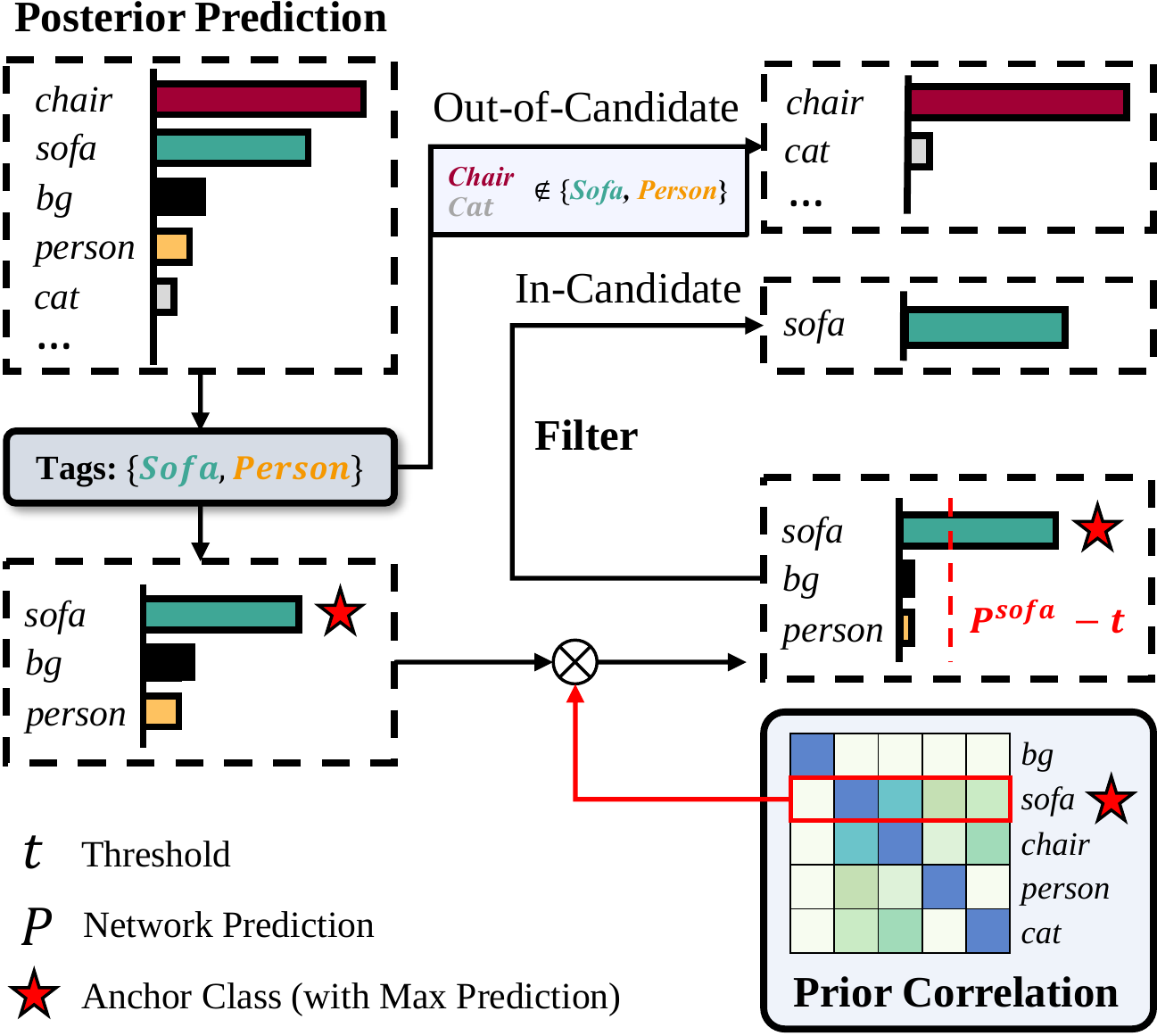}
\caption{\textbf{Adaptive Split.} The OC classes group are selected out according to the contradiction with the tag labels. For IC classes group, we firstly select anchor class $ \mathcal{A} $ (i.e, max prediction class). Specifically, the anchor class is "\textit{sofa}". Then we use prior correlation of anchor class $ \mathcal{M}_{\mathcal{A}} $ to modulate the prediction score of classes in tags $ S $. Finally, those classes in tags whose prediction score is lower than threshold ($P^{\mathcal{A}} - t $) are filtered.}
\label{fig:ic_oc_split}
\vspace{-15pt}
\end{figure}

\textbf{IC and OC Group Split.} We define two groups: IC group $ G_{ic} $ and OC group $ G_{oc} $. 
We require IC group and OC group to satisfy such a group ranking relationship:
\begin{equation}
\label{rel_ic_oc}
    z^{k} > z^{l} \ \ \ \ \mathrm{s.t.} \  \forall k\in G_{ic}, \forall l\in G_{oc}.
\end{equation}
Intuitively, we can assign those classes which belong to label candidate set with background, i.e., $ k\in S\cup \{bg\} $, as IC group and those classes which do not belong to label candidate set, i.e., $ l\notin S\cup\{bg\} $, as OC group. But the latent correlation between categories is not considered. We first count the prior correlation matrix $ \mathcal{M} $ according to the co-occurrence of different classes:
\begin{equation}
\label{prior_corr_matrix}
    \mathcal{M}_{k,l} = \frac{\sum^{L}_{i=1}\mathbbm{1}_{k\in S_i, l\in S_i}}{L},
\end{equation}
where $ L $ is the number of samples in whole dataset, $ S_i $ denotes image-level tag set of $i$-th sample and $\mathcal{M}_{k,l}$ means the correlation score between $k$-th class and $l$-th class. Then we define anchor class:
\begin{equation}
\label{anchor_class}
    \mathcal{A} = \mathop{\mathrm{argmax}}\limits_{j\in S_i\cup\{bg\}}\left(z^j\right),
\end{equation}
where $\mathcal{A}$ denotes the class index of anchor class. For OC classes group, it can be easily identified by the contradiction with image-level tags. For IC classes group, we hope to contain as few useless classes as possible while keeping the correct class.
Note that anchor class has a high probability of being the ground truth so that we use anchor class as a ruler to filter out useless classes in image-level tags for building IC classes group. According to the analysis above, we develop an adaptive strategy to split IC and OC Group:
\begin{align}
G_{ic} &= \left\{k|k\in S\cup\{bg\}, P^{A} - P^{k}\times\mathcal{M}_{\mathcal{A},k} < t\right\}, \\
G_{oc} &= \left\{l\ |l\notin S\cup \{bg\}\right\},
\end{align}
where $P$ is the posterior probability prediction from network (i.e., segmentation logits after softmax operation) and $t$ is a threshold for filtering useless classes for IC classes group. To better understand the adaptive split strategy, the split process is illustrated in Fig.~\ref{fig:ic_oc_split}.

\textbf{Rectification Loss.} The core idea of OCR is to rectify those OC pixels and let these pixels have higher activation for IC classes group than OC classes group. To achieve this, we formulate a group ranking problem and expect Eq.~\ref{rel_ic_oc} to hold. The optimization objective can be defined as:
\begin{equation}
\label{opt_ic_oc}
    \mathop{\mathrm{max}}\limits_{l \in G_{oc}}z^{l} < \mathop{\mathrm{min}}\limits_{k\in G_{ic}}z^{k},
\end{equation}
where $ z^{k}_{i} $ denotes the classes in the IC group and $ z^{l}_{i} $ denotes the classes in the OC group, which can be also written as $\mathop{\mathrm{max}}\limits_{l \in G_{oc}}z^{l} - \mathop{\mathrm{min}}\limits_{k\in G_{ic}}z^{k} < 0$. The Eq.~\ref{opt_ic_oc} means we force the minimum of the OC pixels logits for the categories in IC group $ G_{ic} $ to be larger than the maximum of the OC pixels logits for the categories in OC group $ G_{oc} $. To this purpose, we refer to some wonderful works in metric learning~\cite{sun2020circle,wang2017normface,wang2018cosface,schroff2015facenet,hermans2017defense} and semi-supervised learning~\cite{qiao2022pcr} for making Eq.~\ref{opt_ic_oc} as a loss function:
\begin{equation}
\label{loss_opt_ic_oc_v1}
    \mathcal{L}_{rec} = \mathop{\mathrm{max}}\limits_{k \in G_{oc}}z^{l} - \mathop{\mathrm{min}}\limits_{k \in G_{ic}}z^k + \Delta,
\end{equation}
where $ \Delta $ is a value margin that can make the loss function more scalable. But this loss function can not directly attend the gradient backward process of gradient descent optimization since the $ \mathop{\mathrm{max}} $ and $ \mathop{\mathrm{min}} $ functions are globally non-differentiable. To make equation $ \mathcal{L}_{rec} $ differentiable, we adopt smooth approximation from previous research~\cite{nielsen2016guaranteed}:
\begin{equation}
\label{smooth_appro_max}
\begin{aligned}
    \mathrm{max}(z^1, z^2, ..., z^n) &\approx \mathrm{log}(\sum^{n}_{i=1}\mathrm{e}^{z^i}). \\
\end{aligned}
\end{equation}
Based on the above functional approximation, we derive our rectification loss $ \mathcal{L}_{rec} $ for single pixel as:
\begin{align}
\label{loss_opt_ic_oc_v2}
    \mathcal{L}_{rec} &= \mathop{\mathrm{max}}\limits_{k \in G_{oc}}z^{k} - \mathop{\mathrm{min}}\limits_{l \in G_{ic}}z^l + \Delta \\
                 &= \mathop{\mathrm{max}}\limits_{k \in G_{oc}}z^{k} + \mathop{\mathrm{max}}\limits_{l \in G_{ic}}(-z^l) + \Delta \\
                 &= \sum_{l\in G_{oc}}\mathrm{e}^{z^l + \Delta} \times \sum_{k\in G_{ic}}\mathrm{e}^{-z^k}
\end{align}
To avoid excessive optimization, we use \textit{ReLU} function to rectify loss: 
\begin{equation}
\label{loss_opt_ic_oc_v3}
    \mathcal{L}_{rec} = ReLU\left(\sum_{k\in G_{ic}}\mathrm{e}^{-z^k} \times \sum_{l\in G_{oc}}\mathrm{e}^{z^l + \Delta}\right)
\end{equation}
We convert \textit{ReLU} to its smooth approximation~\cite{glorot2011deep} for acquiring a gradient friendly loss:
\begin{equation}
\label{smooth_appro_relu}
\begin{aligned}
    ReLU(z) = \mathrm{max}(z, 0) &\approx \mathrm{log}(1 + \mathrm{e}^z)
\end{aligned}
\end{equation}
The final formula of rectification loss is
\begin{equation}
\label{rec_loss}
\begin{aligned}
    \mathcal{L}_{rec} = m_{oc} \mathrm{log}\left[1 + \sum_{k\in G_{ic}}\mathrm{e}^{-z^k} \times \sum_{l\in G_{oc}}\mathrm{e}^{z^l + \Delta}\right]
\end{aligned}
\end{equation}
where $ m_{oc} $ is the OC pixel selection mask.

\section{Experiments}

\begin{table}[t]
\small
\centering
\tablestyle{1pt}{1.0}
\begin{tabular}{z{80}|x{50}|x{40}x{40}}
\thickhline
\multicolumn{1}{c|}{\multirow{2}{*}{Method}} & \multirow{2}{*}{Seg} & \multicolumn{2}{c}{mIoU (\%)} \\
& &  \textit{val} & \textit{test}  \\ 
\hline
SS-WSSS~\pub{CVPR20}{~\cite{araslanov2020single}} & V2-Res38 & {62.7} & {64.3} \\
$^*$OAA$^+$~\pub{ICCV19}{~\cite{jiang2019integral}} & V2-Res101 & {65.2} & {66.4} \\
$^\ddagger$BES~\pub{ECCV20}{~\cite{chen2020weakly}} & V2-Res101 & {65.7} & {66.6} \\
$^*$MCIS~\pub{ECCV20}{~\cite{sun2020mining}} & V2-Res101 & {66.2} & {66.9}\\
$^*$ICD~\pub{CVPR20}{~\cite{fan2020learning}} & V2-Res101 & {67.8} & {68.0} \\
$^\ddagger$AdvCAM~\pub{CVPR21}{~\cite{lee2021anti}} & V2-Res101 & {68.1} & {68.0} \\
$^*$NSROM~\pub{CVPR21}{~\cite{yao2021non}} & V2-Res101 & {68.3} & {68.5} \\
$^*$GroupWSSS~\pub{AAAI21}{~\cite{li2021group}} & V2-Res101 & {68.7} & {69.0} \\
$^\dagger$EDAM~\pub{CVPR21}{~\cite{wu2021embedded}} & V2-Res101 & {70.9} & {70.6} \\ 
$^\ddagger$AMR~\pub{AAAI22}{~\cite{qin2022activation}} & V2-Res101 & {68.8} & {69.1} \\
URN~\pub{AAAI22}{~\cite{li2022uncertainty}} & V2-Res101 & {69.5} & {69.7} \\
AFA~\pub{CVPR22}{~\cite{ru2022learning}}    & MiT-B1~\cite{xie2021segformer} & {66.0} & {66.3} \\
\hline
$^\dagger$AffinityNet~\pub{CVPR18}{~\cite{ahn2018learning}} & V1-Res38 & 61.7 & 63.7 \\
$^\dagger$SSDD~\pub{ICCV19}{~\cite{shimoda2019self}} & V1-Res38 & {64.9}& {65.5} \\
$^\dagger$SEAM~\pub{CVPR20}{~\cite{wang2020self}} & V1-Res38 & {64.5} & {65.7} \\
$^\dagger$CONTA~\pub{NeurIPS20}{~\cite{zhang2020causal}} & V1-Res38 & {66.1} & {66.7} \\
$^\ddagger$CDA~\pub{ICCV21}{~\cite{su2021context}} & V1-Res38 & {66.1} & {66.8} \\
$^\dagger$CPN~\pub{ICCV21}{~\cite{zhang2021complementary}} & V1-Res38 & {67.8} & {68.5} \\
$^\ddagger$OC-CSE~\pub{ICCV21}{~\cite{kweon2021unlocking}} & V1-Res38 & {68.4} & {64.2} \\
MCTformer~\pub{CVPR22}{~\cite{xu2022multi}}    & V1-Res38  & {71.9} & {71.6} \\

\hline
		
\rowcolor{aliceblue} $^*$AffinityNet & V1-Res38 & {61.7} & {63.7} \\
\rowcolor{aliceblue} $^*${\textbf{OCR}}\baseline{+AffinityNet} & V1-Res38 & \bslimp{{64.9}}{3.2} & \bslimp{{65.2}}{1.5} \\  \hdashline
		
\rowcolor{aliceblue} $^\dagger$SEAM & V1-Res38 & {64.5} & {65.7} \\
\rowcolor{aliceblue} $^\dagger${\textbf{OCR}}\baseline{+SEAM}  & V1-Res38 & \bslimp{{67.8}}{3.3} & \bslimp{{68.4}}{2.7}\\  

\hline
		
\rowcolor{aliceblue} MCTformer & V1-Res38 & {71.9} & {71.6} \\
\rowcolor{aliceblue} {\textbf{OCR}}\baseline{+MCTformer} & V1-Res38 & \bslimp{{72.7}}{0.8} & \bslimp{{72.0}}{0.4}\\  
\thickhline
\end{tabular}
\caption{\small\textbf{Main Results on Pascal VOC}~\cite{everingham2010pascal} \textit{val} and \textit{test} split. Seg denotes the segmentation networks used by models. For instance, V1-Res38 denotes Deeplabv1~\cite{chen2014semantic} based on ResNet38 and V2-Res101 is DeeplabV2~\cite{chen2017deeplab} based on ResNet101. $^*$, $^\dagger$ and $^\ddagger$ denote models using VGG16, ResNet38 or ResNet50 as the classification network backbone. Besides, segmentation and classification of SS-WSSS~\cite{araslanov2020single}, URN~\cite{li2022uncertainty} and AFA~\cite{ru2022learning} share same network. The classification network of MCTformer is DeiT-S\cite{touvron2021training}.}
\vspace{-15pt}
\captionsetup{font=small}
\label{tab:main_res_voc}
\end{table}

\subsection{Experimental Settings}
\par\textbf{Datasets.} We evaluate our method on two datasets, \ie, PASCAL VOC 2012~\cite{everingham2010pascal} and MS COCO 2014~\cite{lin2014microsoft}. \textbf{PASCAL VOC} has 1,464, 1,449, and 1,456 images for training (train), validation (val) and test sets, respectively. It has 20 object classes and one background class. Following the common practice of prior works~\cite{chang2020weakly,wang2020self,lee2021anti,su2021context,zhang2021complementary,xu2021leveraging}, an augmented set of 10,582 images, with additional data from~\cite{hariharan2011semantic}, was used for training. Furthermore, \textbf{MS COCO} consists of 80 object classes and one background class whose training and validation sets contain 82,081 and 40,137 images, respectively. Following \cite{lee2021railroad,choe2020attention}, we remove images without target classes and adopt the ground-truth labels of COCO stuff\cite{caesar2018coco}. 

\par\textbf{Evaluation protocol.}
To be consistent with previous works~\cite{lee2021anti,xu2022multi}, we adopt the mean Intersection-over-Union (mIoU) to evaluate the semantic segmentation performance on the \textit{val} set of two datasets. The semantic segmentation results on the PASCAL VOC \textit{test} set are acquired from the official PASCAL VOC online evaluation server.

\par\textbf{Implementation details.}
In line with our baseline methods (\ie, AffinityNet~\cite{ahn2018learning}, SEAM~\cite{wang2020self}, MCTformer~\cite{xu2022multi}), we choose DeepLab-LargeFOV (V1)~\cite{chen2014semantic} based on ResNet38~\cite{wu2019wider} backbone network whose output stride is 8 as our segmentation network. All of the backbone networks are pre-trained on ImageNet~\cite{deng2009imagenet}. Note that we calculate a prior correlation score matrix $ \mathcal{M} $ by counting the co-occurrence between different classes, the detailed scores are available in the appendix. The threshold $ t $ is set to 0.2. At train time, we use SGD whose momentum and weight decay are 0.9 and 5e-4 as our optimizer. The initial learning rate is 1e-3 which is multiplied by an exponential decay factor during training process. We train our models for 30 epochs with a batch size of 16. For data augmentation, the training images are randomly rescaled with a scale ratio from 0.7 to 1.3 and then are cropped to $ 321 \times 321 $. At test time, we use test-time augmentation and DenseCRFs with the hyper-parameters suggested in~\cite{chen2014semantic} for post-processing.

\begin{table}[t]
\small
\centering
\tablestyle{1pt}{1.0}
\begin{tabular}{z{80}|x{50}|x{50}|x{40}}
\thickhline
\multicolumn{1}{c|}{\multirow{2}{*}{Method}} & \multirow{2}{*}{Cls} & \multirow{2}{*}{Seg} & mIoU (\%) \\
& &  & \textit{val}  \\ 
\hline
IRNet~\pub{CVPR19}{~\cite{ahn2019weakly}}         & Res50  & V2-Res50  & 32.6 \\ 
IAL~\pub{IJCV20}{~\cite{wang2020weakly}}          & Res38  & V2-VGG16  & 27.7 \\
Luo \etal~\pub{AAAI20}{~\cite{luo2020learning}}   & Res101 & V2-VGG16  & 29.9 \\
Group\-WSSS~\pub{AAAI21}{~\cite{li2021group}}     & VGG16  & V2-VGG16  & 28.4 \\
URN~\pub{AAAI22}{~\cite{li2022uncertainty}}       & Res101 & V2-Res101 & 40.7 \\
AFA~\pub{CVPR22}{~\cite{ru2022learning}}            & MiT-B1 & MiT-B1    & 38.9 \\

\hline

$^*$AffinityNet~\pub{CVPR18}{~\cite{ahn2018learning}} & Res38  & V1-Res38  & 29.5 \\
SEAM~\pub{CVPR20}{~\cite{wang2020self}}           & Res38  & V1-Res38  & 31.9 \\
CONTA~\pub{NeurIPS20}{~\cite{zhang2020causal}}    & Res38  & V1-Res38  & 32.8 \\
OC-CSE~\pub{ICCV21}{~\cite{kweon2021unlocking}}   & Res38  & V1-Res38  & 36.4 \\
CDA~\pub{ICCV21}{~\cite{su2021context}}           & Res38  & V1-Res38  & 33.2 \\
MCTformer~\pub{CVPR22}{~\cite{xu2022multi}}       & Res38  & V1-Res38  & 42.0 \\

\hline
		
\rowcolor{aliceblue} $^*$AffinityNet & Res38 & V1-Res38 & {29.5} \\
\rowcolor{aliceblue} {\textbf{OCR}}\baseline{+AffinityNet} & Res38 & V1-Res38 & \bslimp{{30.5}}{1.0} \\  \hdashline
		
\rowcolor{aliceblue} SEAM & Res38 & V1-Res38 & {31.9} \\
\rowcolor{aliceblue} {\textbf{OCR}}\baseline{+SEAM}  & Res38 & V1-Res38 & \bslimp{{33.2}}{1.3} \\

\hline
		
\rowcolor{aliceblue} MCTformer & DeiT-S & V1-Res38 & {42.0} \\
\rowcolor{aliceblue} {\textbf{OCR}}\baseline{+MCTformer} & DeiT-S & V1-Res38 & \bslimp{{42.5}}{0.5} \\ 
\thickhline
\end{tabular}
\caption{\small\textbf{Main Results on MS COCO}~\cite{lin2014microsoft} \textit{val} split. Cls and Seg denote the classification backbone and segmentation network used by models, respectively. Note that AffinityNet doesn't provide official evaluation results on MS COCO dataset, the results of AffinityNet~\cite{ahn2018learning}(*) are implemented by us.}
\vspace{-15pt}
\captionsetup{font=small}
\label{tab:main_res_coco}
\end{table}

\subsection{Main Results}

For calibrating the training settings of segmentation network, we mainly adopt our OCR into baseline methods (e.g., AffinityNet, SEAM and MCTformer) which use ResNet38 based deeplabv1 as segmentation network. 

\par\textbf{Pascal VOC.} Tab.~\ref{tab:main_res_voc} provides the comparison results of our OCR against representative methods on Pascal VOC \textit{val} split and \textit{test} split. As shown in Tab.~\ref{tab:main_res_voc}, our OCR consistently increases the performance of baseline methods (e.g., AffinityNet, SEAM and MCTformer). Specifically, our OCR can not only improve AffintityNet, SEAM and MCTformer by 3.2\%, 3.3\% and 0.8\% on Pascal VOC \textit{val} split but also improve AffinityNet, SEAM and MCTformer by 1.5\%, 2.7\% and 0.4\% on Pascal VOC \textit{test} split. Besides, OCR with MCTformer sets a new State-Of-The-Art.

\par\textbf{MS COCO.} Tab.~\ref{tab:main_res_coco} reports the results of our OCR compared to previous methods on MS COCO \textit{val} split. MS COCO is a more challenging dataset than Pascal VOC because it has more semantic categories and the OC phenomenon is more likely to occur, where our OCR still improves some previous representative methods (AffinityNet, SEAM and MCTformer) by 1.0\%, 1.3\% and 0.5\%. Same as Pascal VOC, we also build a new State-Of-The-Art by adopting OCR into MCTformer.

\subsection{Quantitative Analysis}

\par\textbf{Extra Computation Cost.} Our OCR is used to boost the training progress of segmentation network and can be removed when evaluating segmentation network. In order to check if it introduces heavy computation overhead when training segmentation network, we evaluate OCR based on SEAM with different image crop size. In Tab.~\ref{tab:comput_overhead}, OCR improve performance (mIoU (\%)) by 3.2$\sim$3.7\% with only 0.56$\sim$1.18 min./epoch additional training time, which shows the effectiveness and efficiency of our OCR.

\begin{figure}[t]
\centering
\includegraphics[width=0.45\textwidth]{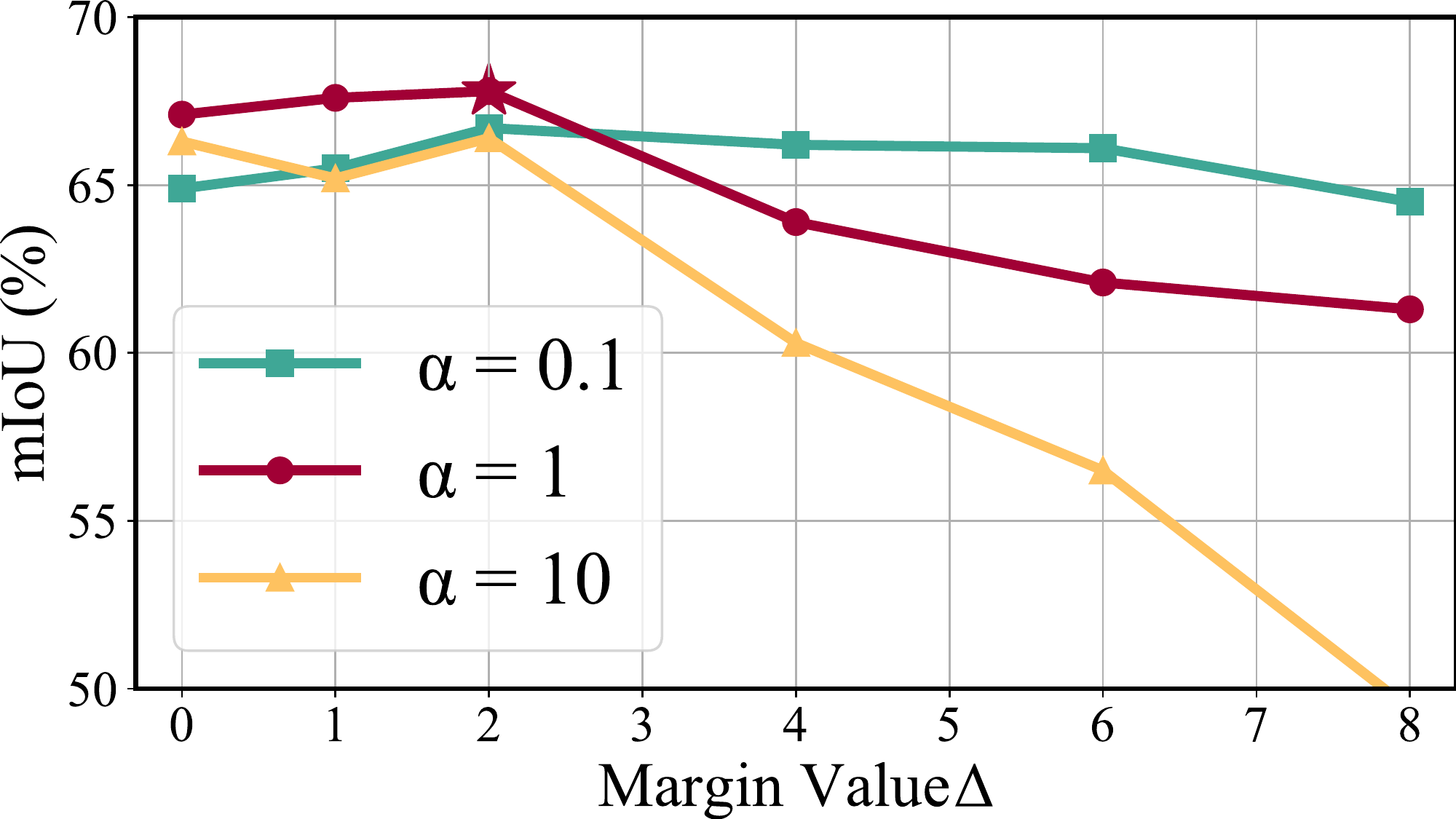}
\caption{The performance (mIoU (\%)) influence of the margin value $ \Delta $ and the loss coefficient $ \alpha $ for penalizing loss $ L_{penalize} $. This study is based on SEAM and the results are evaluated on Pascal VOC \textit{val} split. The best parameter selection of $ \alpha $ and $ \Delta $ are 1 and 2, respectively.}
\label{fig:abl_margin_alpha}
\end{figure}

\newcommand{\improve}[1]{
	\fontsize{7.5pt}{1em}\selectfont\color{purple}{$\uparrow$ \textbf{#1}}
}

\begin{table}[t]
\small
\centering
\begin{tabular}{lccc}
\toprule 
Method & Size & mIoU (\%) & Train Speed \\
\midrule
\multirow{3}{*}{SEAM (\color{purple}{\textbf{w/ OCR}})} & $ 256^2 $  & 63.6 ({\improve{3.2}}) & 14.38 ({\improve{0.56}})   \\
& $ 321^2 $ & 64.5 ({\improve{3.3}}) & 23.32 ({\improve{0.83}})   \\
& $ 448^2 $ & 64.7 ({\improve{3.7}}) & 42.29 ({\improve{1.18}})   \\
\bottomrule
\end{tabular}
\caption{Inference performance (mIoU (\%)) and training speed (min./epoch) of OCR based on SEAM. The size is the crop size when preprocessing images. All of the results are evaluated on Pascal VOC \textit{val} split.}
\vspace{-15pt}
\label{tab:comput_overhead}
\end{table}

\textbf{Different Value Margin.} The margin value $\Delta$ can control prediction differences between IC and OC groups. In other words, it can locally control the intensity of rectification. In order to check the influence of rectification intensity, we conduct extensive experiments on margin $ \Delta $. As shown in Fig.~\ref{fig:abl_margin_alpha}, the overall best choice of $ \Delta $ is 2. When placing a larger value for the margin $ \Delta $, excessive rectification degrades the performance of segmentation network. When placing a smaller value for margin $ \Delta $, insufficient rectification provides suboptimal performance.

\textbf{Loss Coefficient.} We use a loss coefficient $ \alpha $ to globally control the rectification intensity of rectification loss $ L_{rec} $. In Fig.~\ref{fig:abl_margin_alpha}, we conduct empirical experiments to check the influence of the loss coefficient. When we adjust the loss coefficient $ \alpha $ to evaluate segmentation network, we find that the overall optimal choice of $ \alpha $ is 1.0.

\begin{figure}
\centering
\begin{subfigure}{0.495\linewidth}
\includegraphics[width=1\textwidth]{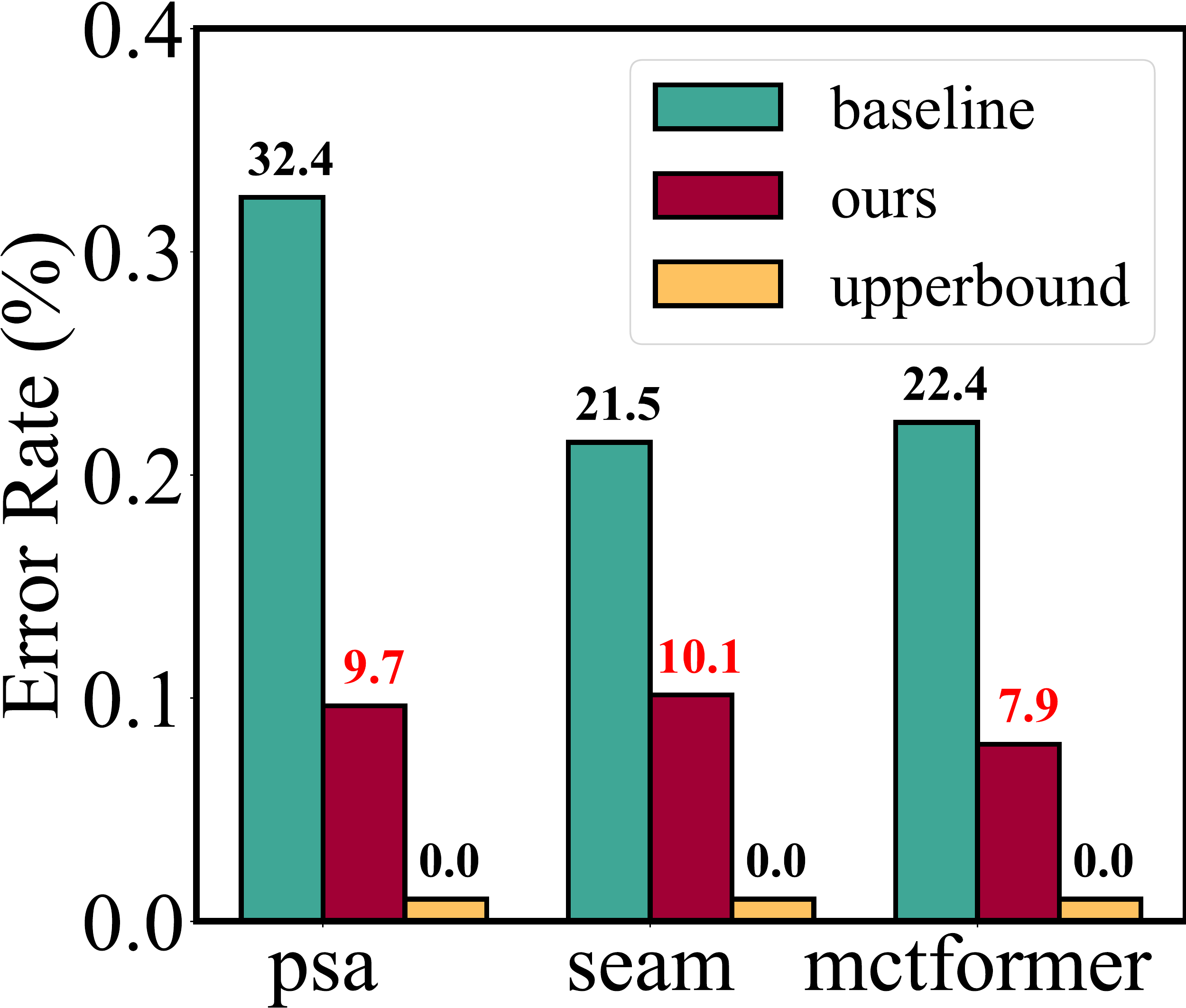}
\caption{Error Rate}
\label{fig:era_error_rate}
\end{subfigure}
\hfill
\begin{subfigure}{0.495\linewidth}
\includegraphics[width=1\textwidth]{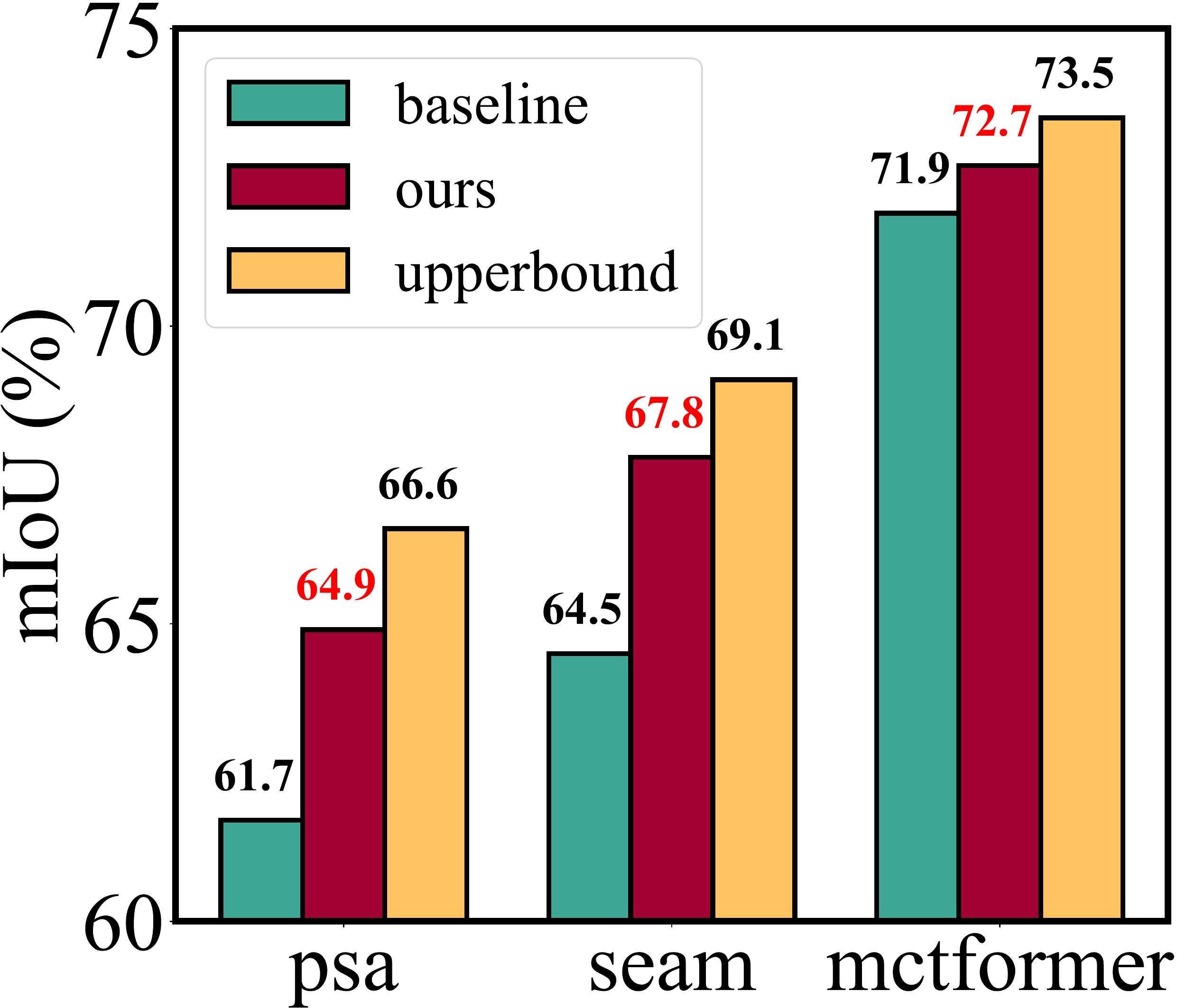}
\caption{Performance}
\label{fig:era_performance}
\end{subfigure}
\caption{Effectiveness analysis of our proposed method. (a). The rate of images with Out-Of-Candidate error on Pascal VOC 2012 \textit{val} split. (b). The mIoU metric on Pascal VOC 2012 \textit{val} split. We choose SEAM as our baseline in this experiment.}
\label{fig:era}
\end{figure}

\newcommand{\increase}[1]{
	\fontsize{7.5pt}{1em}\selectfont\color{purple}{$\uparrow$ \textbf{#1}}
}
\newcommand{\decrease}[1]{
	\fontsize{7.5pt}{1em}\selectfont\color{gray!48}{$\downarrow$ \textbf{#1}}
}

\begin{table}[t]
\small
\centering
\tablestyle{1pt}{1.0}
\begin{minipage}{0.28\textwidth}
\renewcommand{\arraystretch}{1.3}
\centering
\begin{subtable}[t]{1.0\textwidth}
    \begin{tabular}{z{57}|x{25}|x{42}}
    \thickhline
    \multicolumn{1}{c|}{$ G_{ic} $} & $ G_{oc} $ & mIoU \\
    \hline
    None                           & None             & 64.5   \\
    $ \mathrm{all}\left(S \cup \{bg\}\right) $ & $ \overline{S} $ & 63.4~({\decrease{1.1}})   \\
    $ \mathrm{max}\left(S \cup \{bg\}\right) $ & $ \overline{S} $ & 67.2~({\increase{2.7}})   \\
    $ \mathrm{ada}\left(S \cup \{bg\}\right) $ & $ \overline{S} $ & 67.8~({\increase{3.3}})   \\
    \thickhline
    \end{tabular}
    \caption{IC/OC group split.}
    \label{tab:ic_oc_gp_split}
\end{subtable}
\end{minipage}
\begin{minipage}{0.19\textwidth}
\renewcommand{\arraystretch}{1.3}
\centering
\begin{subtable}[t]{1.0\textwidth}
    \begin{tabular}{z{40}|x{45}}
        \thickhline
        \multicolumn{1}{c|}{Rec. Pixels} & mIoU \\
        \hline
        None    & 64.5   \\
        IC      & 64.7~({\increase{0.2}})   \\
        OC      & 67.8~({\increase{3.3}})   \\ 
        ALL     & 66.9~({\increase{2.4}})   \\
        \thickhline
    \end{tabular}
    \caption{Rectified pixels select.}
    \label{tab:rec_pix_sel}
\end{subtable}
\end{minipage}
\vspace{-5pt}
\caption{Ablation study of (a). IC/OC group split strategy; (b). Rectified pixels select strategy.
“Rec. Pixels” means those pixels which will be rectified by our OCR. $ \mathrm{max}\left(\cdot\right) $ denotes select the anchor class which has max prediction score in IC classes group. $ \mathrm{ada}\left(\cdot\right) $ denotes selecting class by considering prior and posterior correlation information. These experiments are based on SEAM.}
\vspace{-15pt}
\end{table}

\begin{figure*}[t]
\centering
\includegraphics[width=1.0\textwidth]{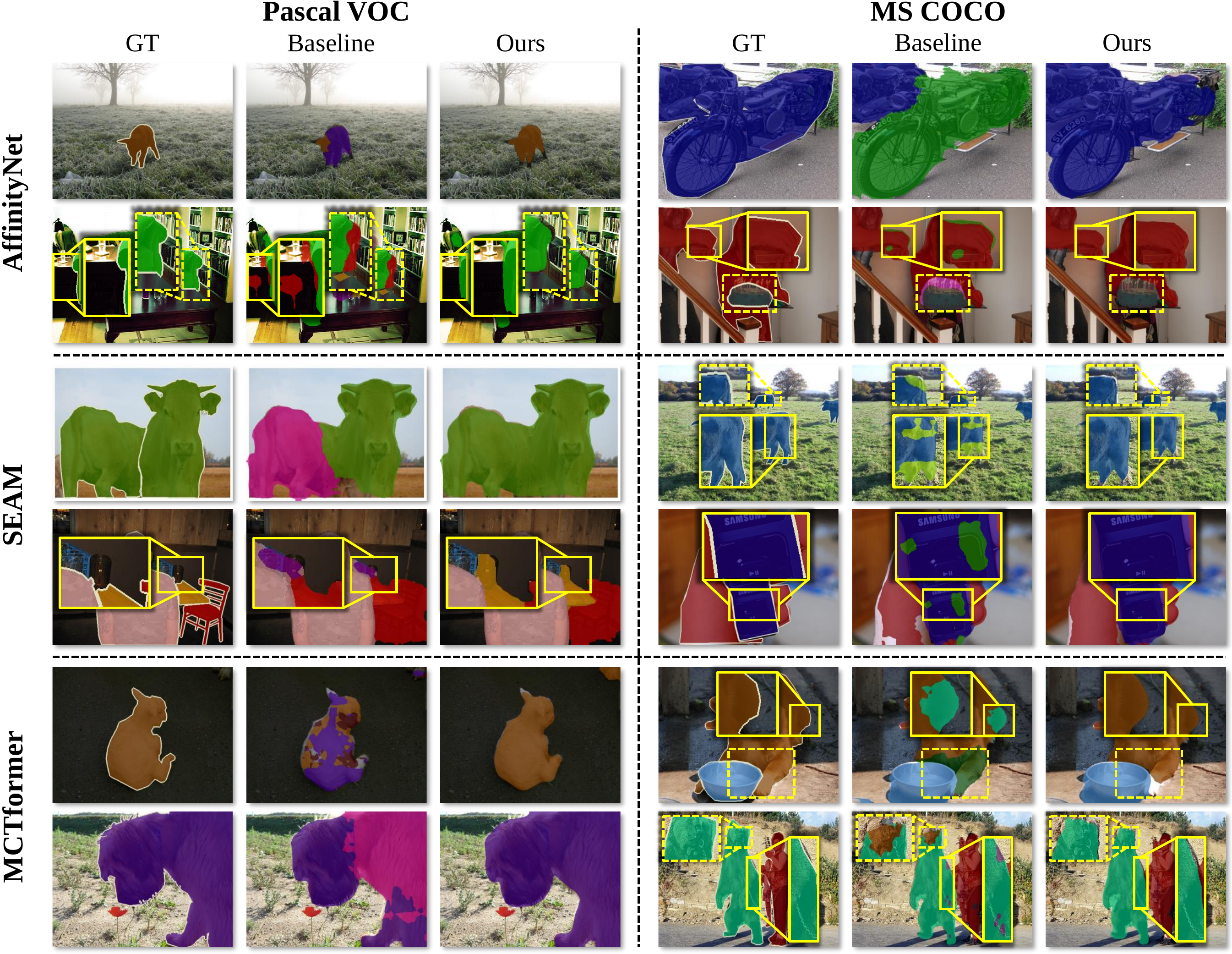}
\caption{\textbf{Qualitative segmentation results comparison} between baseline methods and our methods. Prediction results from baseline methods usually contain Out-Of-Candidate pixels on both Pascal VOC and MS COCO dataset (2nd and 5th columns). Our methods (OCR) can fix these pixels to correct class (3rd and 6th columns). \textbf{\textcolor{myyellow}{Yellow}} boxes are used to highlight the effect of our method.}
\label{fig:qualitative_analysis}
\vspace{-9pt}
\end{figure*}

\textbf{Different Group Split Strategy.} As mentioned in Sec~\ref{subsec:ocr}, IC classes group is just the candidate set of correct classes and is not equivalent to the ground truth of OC pixels. So we have to filter out some useless classes in IC classes group. We compare this adaptive strategy $ ada(\cdot) $ to two brute force strategies: 1). $ \mathrm{all}\left(\cdot\right) $ which means selecting all tag labels into IC classes group; 2). $ \mathrm{max}\left(\cdot\right) $ which means only selecting anchor class (i.e., the class with max prediction score) into IC classes group. In Tab.~\ref{tab:ic_oc_gp_split}, we can find that the OCR with $ \mathrm{all}\left(\cdot\right) $ strategy decreases the baseline performance by 1.1\% mIoU which shows that useless classes in IC classes group cause negative effect. In contrast, we find that $ \mathrm{ada}\left(\cdot\right) $ and $ \mathrm{max}\left(\cdot\right) $ strategies can improve the performance of baseline by 2.7\% and 3.3\% mIoU. The comparison above tells us that it is necessary to remove useless classes in IC classes group. Besides, the $ \mathrm{ada}\left(\cdot\right) $ group split strategy is more effective than the $ \mathrm{max}\left(\cdot\right) $ strategy, which means the anchor class is not always the correct class of OC pixels and our adaptive strategy can keep the correct class of OC pixels better than the $ \mathrm{max}\left(\cdot\right) $ strategy.

\textbf{Different Pixel Selection Strategy.} Pixel selection strategy controls the operation domain of OCR. In Tab.~\ref{tab:rec_pix_sel}, we find that if we only rectify IC pixels, the OCR only brings a negligible $ 0.2\% $ mIoU improvement for baseline. If we rectify all of the pixels in an image, the OCR improves the performance of the baseline by 2.4\% which is lower than only rectifying the OC pixels. So using OCR to extra rectify IC pixels is unnecessary and may cause a suboptimal effect for rectifying OC pixels. Only using OCR to rectify OC pixels achieve optimal improvement for the baseline. 

\textbf{Error Rate Analysis.} In order to verify if the OCR can suppress the occurrence of OC pixels, we count the proportion of OC error predictions of baseline and our method on Pascal VOC \textit{val} split. As shown in Fig.~\ref{fig:era_error_rate}, our proposed OCR significantly reduces the proportion of OC error for the baseline methods (i.e., AffinityNet, SEAM and MCTformer) from 32.4\%, 21.5\% and 22.4\% to 9.7\%, 10.1\% and 7.9\%. With the reduction of OC error, the performance of baseline models is synchronously improved from 61.7\%, 64.5\% and 71.9\% mIoU to 64.9\%, 67.8\% and 72.9\% mIoU. 

\subsection{Qualitative Analysis}
 Fig.~\ref{fig:qualitative_analysis} depicts qualitative comparisons of our method with baseline methods against baseline methods (i.e., AffinityNet, SEAM and MCTformer) over representative examples on both Pascal VOC and MS COCO datasets. We can clearly show that those OC pixels occurring on baseline predictions are rectified after adopting our OCR whether it is a simple scenario that only contains a single object (e.g., the first row and the last three columns) or a complex scenario that contains multiple objects (e.g., the second row and the first three columns).

\section{Conclusion}
In this paper, we observe that previous WSSS methods usually output pixels whose semantic categories are in contradiction with image-level candidate tags. Then we propose some new concepts (OC/IC) to describe this special type of segmentation error. To tackle these errors, we propose Out-of-Candidate Rectification (OCR). The OCR first defines IC and OC classes group. Then we formulate the relationship between IC and OC groups by a group ranking problem. Finally, we derive a differentiable rectification loss to solve the group ranking problem for suppressing the OC phenomenon. We incorporate OCR with several representative baseline methods for evaluation. The experiments show that our OCR can consistently improve baseline methods on both Pascal VOC and MS COCO datasets, which can demonstrate the effectiveness and generality of our OCR.

{\small
\bibliographystyle{ieee_fullname}
\bibliography{main}
}

\clearpage
\appendix
\onecolumn

\begin{center}
    \Large \bf Appendix
\end{center}
\vspace{15pt}

In this document, we first discuss the limitations of our method~(Sec.~\ref{sup_sec:dis}). Then we provide qualitative analysis of different pixel selection strategies~(\S Sec.~\ref{sup_sec:qual_ana_pix_sel_str}), which is a supplement to quantitative analysis of pixel selection strategies in main paper. Thirdly, the detailed prior correlation information is illustrated and described~(Sec.~\ref{sup_sec:pri_corr_infor}). Finally, pseudo-codes of OCR~(\S Sec.~\ref{sup_sec:pseu_code}) and extra qualitative results~(Sec.~\ref{sup_sec:extra_qual_res}) are provided.

\section{Discussion}
\label{sup_sec:dis}

In this section, we discuss the limitations of our Out-of-Candidate Rectification (OCR). The limitations can be divided into two folds: Firstly, although the OCR is proposed for Out-of-Candidate (OC) phenomenon, the OCR can't drastically solve the OC problem. OC phenomenon is essentially the intrinsic weakness of weakly supervised semantic segmentation and this problem may be intractable under current CAM-based technology. The second fold is that the proposed OCR requires a strong hypothesis, i.e., ``\textit{anchor class which has maximal prediction score is the ground truth class}". However, the hypothesis does not always hold so it is not always possible to rectify OC pixels to ground truth class.

\section{Qualitative Analysis of Pixel Selection Strategy}
\label{sup_sec:qual_ana_pix_sel_str}

\begin{figure}[ht]
\centering
\includegraphics[width=\textwidth]{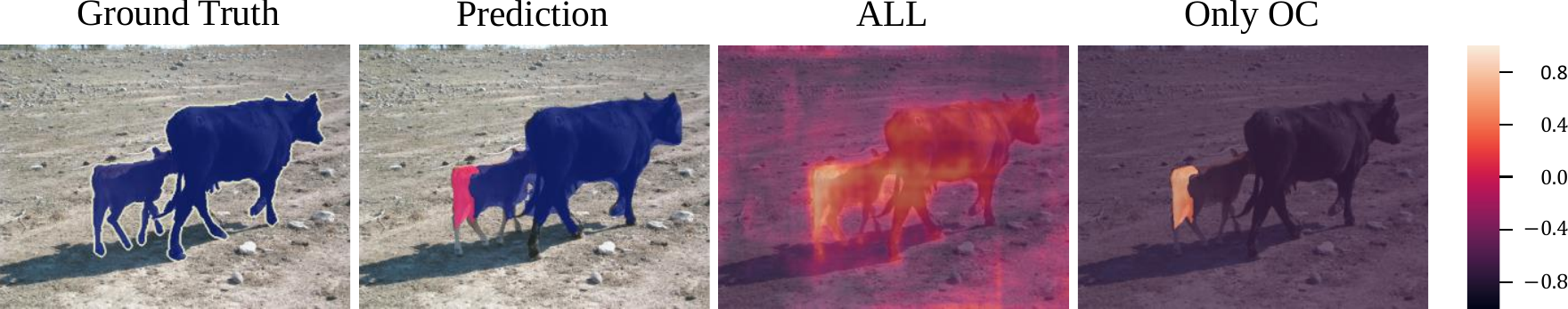}
\caption{\textbf{Loss spatial map visualization of OCR with different pixel selection strategy.} ``ALL" denotes we select all of the pixels to attend the calculation of OCR loss. ``Only OC" means only the OC pixels are selected to attend the calculation of OCR loss. ``ALL" pixel selection strategy provides unnecessary supervision signals to IC pixels, which causes suboptimal performance. ``Only OC" strategy is highly selective to OC pixels, which can efficiently suppress OC error phenomenon for better performance. \textbf{\color{voc_cow}{Blue}} denotes ``\textbf{\textit{\color{voc_cow}{cow}}}" and \textbf{\color{voc_horse}{Pink}} denotes ``\textbf{\textit{\color{voc_horse}{horse}}}".}
\label{fig:loss_map}
\end{figure}

In Fig.~\ref{fig:loss_map}, we provide the spatial visualization of OCR spatial loss map by two types of pixel selection strategies (``ALL" and ``Only OC"). By comparing the first column and second column of Fig.~\ref{fig:loss_map}, we can easily identify those \textbf{\color{voc_horse}{Pink}} pixels as OC pixels. These OC pixels whose ground truth label is ``\textbf{\textit{\color{voc_cow}{cow}}}" are misled to ``\textbf{\textit{\color{voc_horse}{horse}}}". The third column and fourth column of Fig.~\ref{fig:loss_map} are the loss spatial map of OCR with ``ALL" or ``Only OC" pixel selection strategies, respectively. ``ALL" strategy assign different intensity of supervision signals to all of the pixels, which means IC pixels are also forced to attend to the rectification of OCR. Rectification loss of OCR is only used to rectify OC pixels from OC group to IC group. However, IC pixels already belong to IC group so the rectification loss of OCR calculated on IC pixels is not useful for improving results. ``Only OC" strategy only assigns supervision signals to OC pixels, which is reasonable and is in line with the motivation of proposed OCR (correcting OC pixels into IC group).

\section{Prior Correlation Information}
\label{sup_sec:pri_corr_infor}

\begin{figure}[t]
\centering
\begin{subfigure}{1.0\linewidth}
\centering
\includegraphics[width=0.6\textwidth]{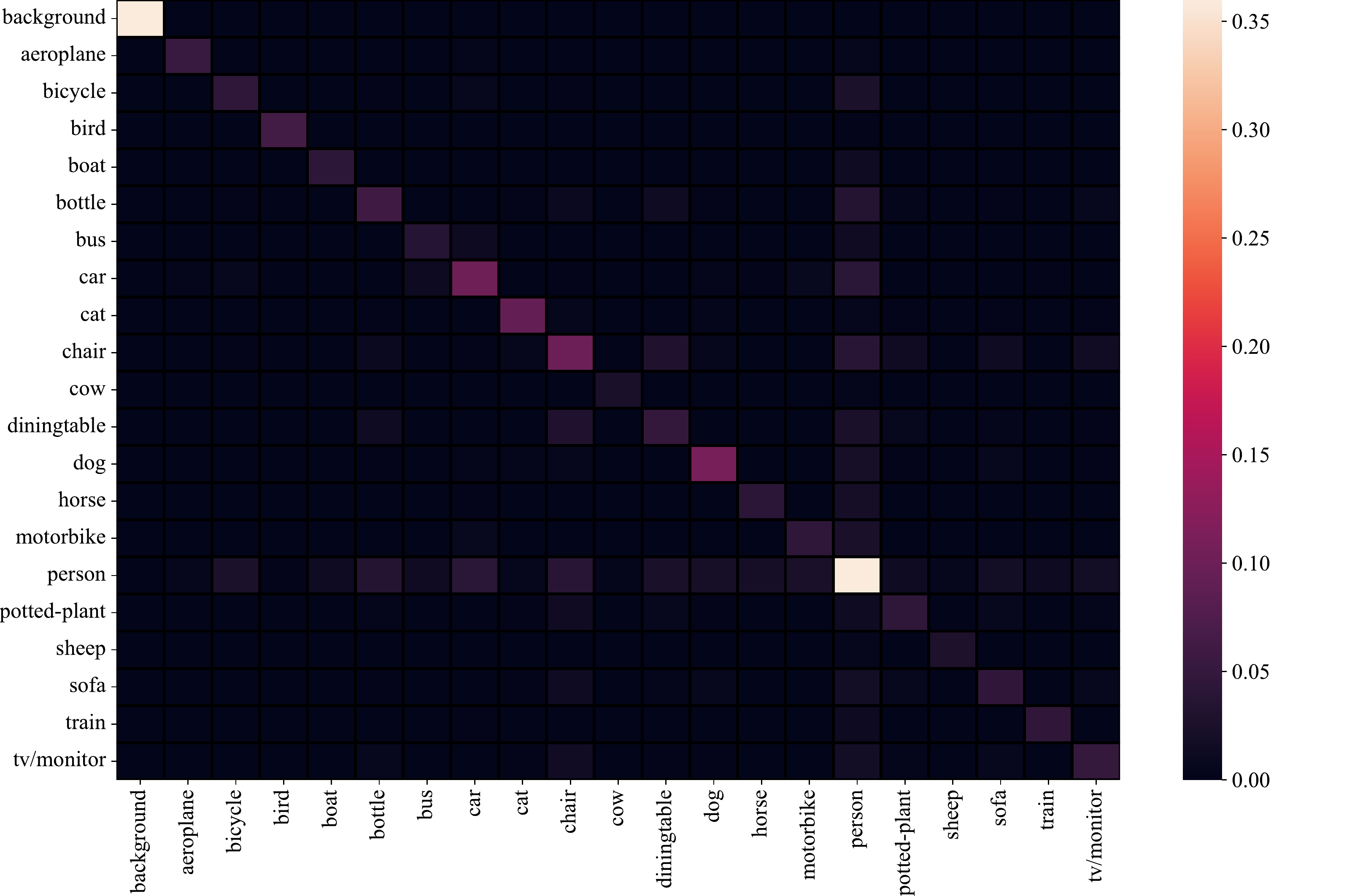}
\caption{Prior Correlation}
\label{fig:pri_corr}
\end{subfigure}
\vfill
\begin{subfigure}{1.0\linewidth}
\centering
\includegraphics[width=0.6\textwidth]{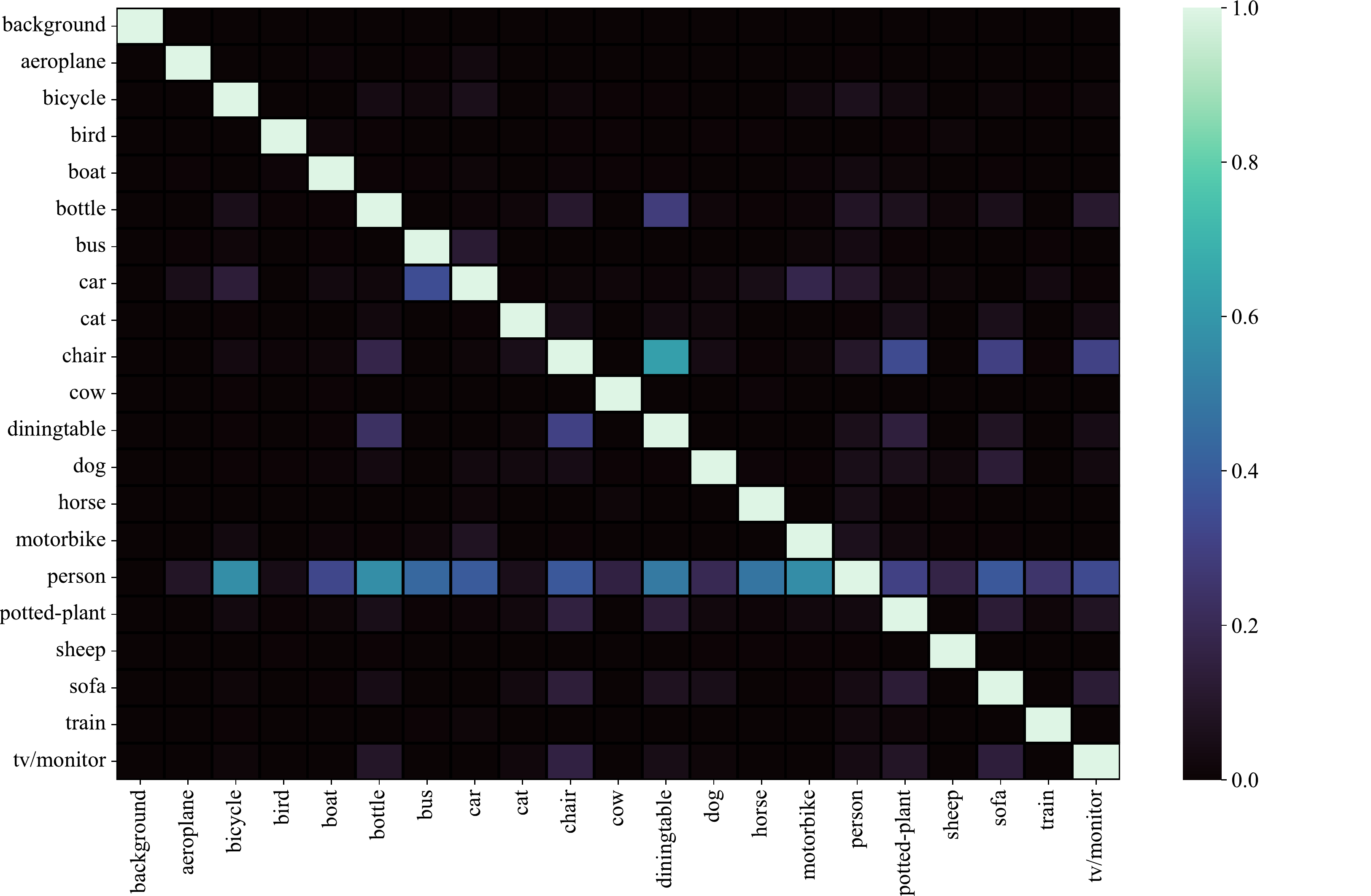}
\caption{Normalized Prior Correlation}
\label{fig:pri_corr_norm}
\end{subfigure}
\caption{\textbf{Visualization of prior correlation matrix.} The original prior correlation matrix calculated by co-occurrence between different classes is shown in (a). For clearly showing the correlation between different classes, we conduct min-max normalization on prior correlation matrix along the horizontal dimension, which is shown in (b).}
\label{fig:pri_corr_total}
\end{figure}

As mentioned in the description of adaptive group split strategy in Out-Of-Candidate Rectification (OCR) Section, we respectively adopt prior and posterior correlation information from co-occurrence of tag labels and network predictions for filtering useless classes in IC group. In Fig.~\ref{fig:pri_corr}, we illustrate the prior correlation matrix of Pascal VOC \textit{train} split. For clearly showing the correlation between different classes, we extra provide normalized prior correlation matrix in Fig.~\ref{fig:pri_corr_norm}. There are some interesting correlation relationship between different classes, e.g., \textit{bus} + \textit{car}, \textit{chair} + \textit{diningtable} and \textit{bottle} + \textit{dingingtable}. The most two special classes are \textit{person} and \textit{background}. The former nearly is highly correlated to nearly all of other classes (except \textit{background}). For latter, \textit{background} is contained in any images of dataset so it has no correlation to other classes.

\section{Pseudo Code}
\label{sup_sec:pseu_code}

In this section, we provide the pseudo-codes of OCR for better grasping our method. We first provide general workflow of OCR in Algo.~\ref{alg:ocr}. OCR comprises three components: Then we describe the details of adaptive IC/OC group split strategy in Algo.~\ref{alg:ada_split}.

\begin{algorithm}[t]
\caption{PyTorch-style Pseudo-code of Out-of-Candidate Rectification (OCR)}
\label{alg:ocr}
\algcomment{
\textbf{Notes}: 
\texttt{sum}($\cdot$), \texttt{exp}($\cdot$) and \texttt{log}($\cdot$) are the functions of pytorch. 
\texttt{[1:]} denotes slice operation which removes the zero element. \texttt{ic\_gp\_build}($\cdot$) and \texttt{oc\_gp\_build}($\cdot$) are introduced in algo.~\ref{alg:ada_split}.
}
\definecolor{codeblue}{rgb}{0.25,0.5,0.5}
\definecolor{codekw}{rgb}{0.85, 0.18, 0.50}
\begin{lstlisting}[language=python]
# im_gt: image-level tag labels [C]
# logits: network output logits [(C + 1),H,W]
# delta: the margin value

# Out-of-Candidate Rectification Loss
def ocr_loss(logits, im_gt):
    # OC pixels selection mask
    clean_logits = logits.detach().clone()
    ## broadcast operation
    clean_logits[1:] = clean_logits[1:] * im_gt
    pred = torch.argmax(logits, dim=0)
    clean_pred = torch.argmax(clean_logits, dim=0)
    ## bool index [H,W]
    oc_mask = (clean_pred != pred)
    
    # Adaptive Group Split
    ic_gp = ic_gp_build(logits, im_gt)
    oc_gp = oc_gp_build(logits, im_gt)
    
    # Rectification Loss
    se_ic = sum(exp(-ic_gp), dim=0)
    se_oc = sum(exp(oc_gp + delta), dim=0)
    l_rec = (oc_mask * log(1 + se_ic * se_oc)).mean()
    
    return l_rec
\end{lstlisting}
\end{algorithm}

\begin{algorithm}[t]
\caption{PyTorch-like Pseudo-code of adaptive IC/OC group split.}
\label{alg:ada_split}
\algcomment{
\textbf{Notes}: \texttt{cat}($\cdot$) is the concatenate operation. \texttt{max}($\cdot$) returns the max value and related indices of tensor. \texttt{nonzero}($\cdot$) is used to acquire the indices of non-zero elements. \texttt{zeros}($\cdot$) is used to create a special shape of tensor filled with 0.}
\definecolor{codeblue}{rgb}{0.25,0.5,0.5}
\definecolor{codekw}{rgb}{0.85, 0.18, 0.50}
\begin{lstlisting}[language=python]
# im_gt: image-level tag labels [C]
# logit: single pixel network output [(C + 1),H,W]
# M: prior correlation matrix [(C + 1),(C + 1)]
# tau: IC group useless classes filter threshold

def get_ic_ids(im_gt):
    # acquire initial IC class ids (tag labels and background)
    return cat([0, nonzero(im_gt) + 1])

def get_oc_ids(im_gt):
    # acquire OC class ids
    return nonzero(1 - im_gt) + 1

# single pixel behavior
def ic_gp_build(logit, im_gt):
    ic_cls_ids = get_ic_ids(im_gt)
    # network output probability
    prob = softmax(logit)
    # 1. Extract initial IC classes by tag labels.
    ic_gp_prob = prob[ic_class_ids]
    ic_filter_mask = zeros(len(ic_class_ids))
    # 2. Anchor Predictions and Classes
    anchor_prob, anchor_id = max(ic_gp_prob)
    ic_filter_mask[anchor_id] = 1
    # 3. utilize correlation for filtering
    cond = (anchor_prob - prob * M[anchor_id]) < tau
    ic_filter_mask[cond] = 1
    
    return (ic_filter_mask * logit)[ic_cls_ids]

# single pixel behavior
def oc_group_build(logit, im_gt):
    oc_cls_ids = get_oc_ids(im_gt)
    return logit[oc_cls_ids]

\end{lstlisting}
\end{algorithm}

\section{Extra Qualitative Results}
\label{sup_sec:extra_qual_res}

\begin{figure*}[t]
\centering
\includegraphics[width=1.0\textwidth]{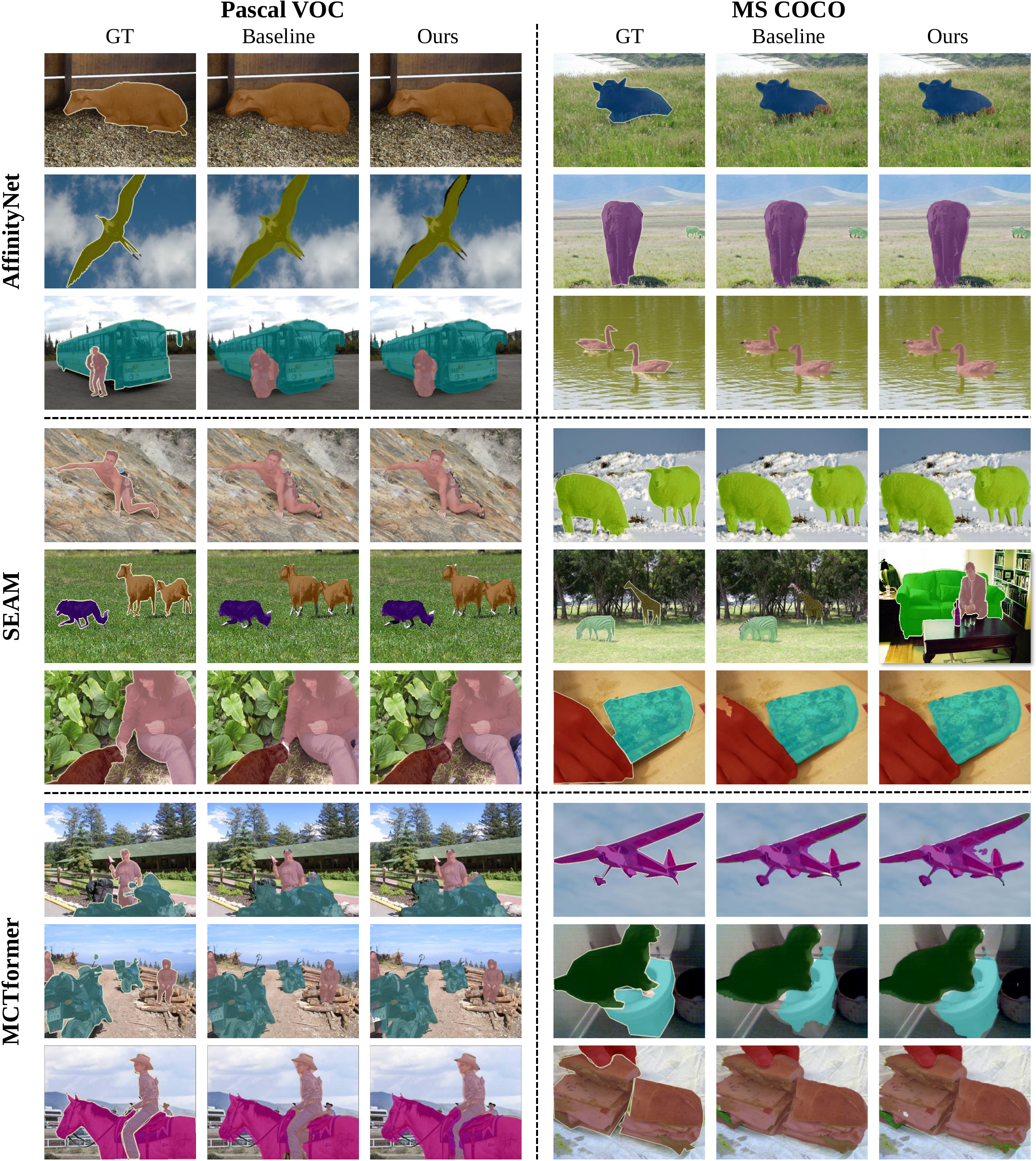}
\caption{\textbf{Extra Qualitative Results.} We extra provide some normal cases of segmentation results for showing regular segmentation performance.}
\label{fig:qual_norm_cases}
\end{figure*}

In main paper, we already provide comparison qualitative results of baseline and our methods for showing the correct effect of OC phenomenon. In this section, some regular cases of segmentation results are shown in Fig.~\ref{fig:qual_norm_cases}.

\end{document}